\documentclass[journal]{IEEEtran}

\usepackage[noadjust]{cite}
\usepackage{pslatex} % -- times instead of computer modern, especially for the plain article class
\usepackage{booktabs}

\usepackage{graphicx}
\usepackage{xcolor}
\usepackage{multirow}

\usepackage{amsmath}
\usepackage{amssymb}
\usepackage{amsfonts} %
\usepackage{calligra} %
\usepackage{mathtools}

\usepackage{amsthm} % for neater definitions

\usepackage{dblfloatfix} %allows floats to be at bottom of page

\usepackage{tikz}
\usepackage{verbatim}
\usetikzlibrary{shapes,backgrounds}
\usetikzlibrary{arrows,fit,automata,positioning,decorations,calc}
\usetikzlibrary{spy}
\usetikzlibrary{matrix,chains,decorations.pathreplacing}

\usepackage{caption}
\usepackage{subcaption} 

\usepackage{comment}
\usepackage{listings}

\usepackage{pgfplots}
\usepackage{pgfplotstable}

\usetikzlibrary{patterns}

\usepackage[percent]{overpic}

\usepackage[english]{babel}
\usepackage{blindtext}

\usepackage{algorithm}
\usepackage{algorithmicx}
\usepackage{algpseudocode}

\algnewcommand\algorithmicforeach{\textbf{for each}}
\algdef{S}[FOR]{ForEach}[1]{\algorithmicforeach\ #1\ \algorithmicdo}

\usepackage{acronym}

\usepackage{listings}
\usepackage{float}
\usepackage{textcomp}

\theoremstyle{definition}
\newtheorem{definition}{Definition}[section] % definition numbers are dependent on theorem numbers
\theoremstyle{example}
\theoremstyle{theorem}

\definecolor{mygreen}{rgb}{0,0.6,0}
\definecolor{mygray}{rgb}{0.5,0.5,0.5}
\definecolor{mymauve}{rgb}{0.58,0,0.82}

\acrodef{WCRT}{Worst Case Reaction Time}
\acrodef{WCET}{Worst-Case Execution Time}
\acrodef{TDM}{Time-Division Multiplexing}
\acrodef{SOT}{Start Of Tick}
\acrodef{EOT}{End Of Tick}

\acrodef{ESS}{Energy Storage System}

\acrodef{CPS}{Cyber-Physical System}
\acrodef{IoT}{Internet of Things}

\acrodef{AI}{Artificial Intelligence}
\acrodef{ANN}{Artificial Neural Network}
\acrodef{CNN}{Convolutional Neural Network}
\acrodef{MLP}{Multi-Layer Perceptron}

\acrodef{SANN}{Synchronous ANN}
\acrodef{STA}{Static Timing Analysis}
\acrodef{OS}{Operating System}
\newcommand{\ignore}[1]{{}}

\newcommand{\squishlist}{
	\begin{list}{$\bullet$}
		{ \setlength{\itemsep}{0pt}
			\setlength{\parsep}{1pt}
			\setlength{\topsep}{1pt}
			\setlength{\partopsep}{0pt}
			\setlength{\leftmargin}{0.9em}
			\setlength{\labelwidth}{1.5em}
			\setlength{\labelsep}{0.4em} } }
	\newcommand{\squishend}{
\end{list}  } 

\newcommand{\sft}[1]{\textup{\sffamily{#1}}}

\newcommand\copyrighttext{%
	\tiny © 2020 IEEE. Personal use of this material is permitted.  Permission from IEEE must be obtained for all other uses, in any current or future media, including reprinting/republishing this material for advertising or promotional purposes, creating new collective works, for resale or redistribution to servers or lists, or reuse of any copyrighted component of this work in other works.}
\newcommand\copyrightnotice{%
	\begin{tikzpicture}[remember picture,overlay]
	\node[anchor=south,yshift=10pt] at (current page.south) {\fbox{\parbox{\dimexpr\textwidth-\fboxsep-\fboxrule\relax}{\copyrighttext}}};
	\end{tikzpicture}%
}

\begin{document}

\title{Designing Neural Networks for Real-Time Systems}

\author{Hammond Pearce,
	    Xin Yang,
	    Partha S. Roop,
	    Marc Katzef,
		T\'{o}rur Biskopst\o{} Str\o{}m%
% make the title area
%
\thanks{H. Pearce, X. Yang, P. S. Roop, and M. Katzef were with the Department
	of Electrical, Computer, and Software Engineering, University of Auckland, New Zealand.}%
\thanks{T. B. Str\o{}m was with the Department of Applied Mathematics and Computer Science, Technical University of Denmark, Denmark.}%
\thanks{Corresponding Email: hammond.pearce@auckland.ac.nz}% <-this % 
%\thanks{Manuscript received April 01, 2020; revised June 30, 2020.}%
}

\maketitle
\copyrightnotice
% As a general rule, do not put math, special symbols or citations
% in the abstract or keywords.

\begin{abstract}
	Artificial Neural Networks (ANNs) are increasingly being used within safety-critical Cyber-Physical Systems (CPSs).
	%They are often co-located with traditional embedded software, and may perform advisory or control-based roles.
	It is important to validate both the timing and functional correctness of these systems.
	However, %while the correctness of CPSs involves meeting both functional and timing properties, 
	most approaches in the literature consider guaranteeing 
	only the functionality of ANN based controllers.
	This issue stems largely from the implementation strategies used within common neural network frameworks --- their underlying source code is often simply unsuitable for formal techniques such as static timing analysis.
	As a result, developers of safety-critical CPS must rely on informal techniques such as measurement based approaches to prove correctness, techniques that provide weak guarantees at best.
		
	In this work we address this challenge. We propose a design pipeline whereby neural networks trained using the popular deep learning framework Keras are compiled to functionally equivalent C code. 
	This C code is restricted to simple constructs that may be analysed by existing static timing analysis tools.
	As a result, if compiled to a suitable time-predictable platform all execution bounds may be statically derived.
	
	To demonstrate the benefits of our approach we execute an ANN trained to drive an autonomous vehicle around a race track.
	We compile the ANN to the Patmos time-predictable controller, and show that we can derive worst case execution timings.
	%To demonstrate the efficacy of our approach a case study 
\end{abstract}

% Note that keywords are not normally used for peerreview papers.
%\begin{IEEEkeywords}
%Keyword, keyword, keyword
%\end{IEEEkeywords}

\IEEEpeerreviewmaketitle

\section{Introduction}
\label{sec:intro}

\begin{figure}[h!]
	\centering
	\includegraphics[scale=0.25]{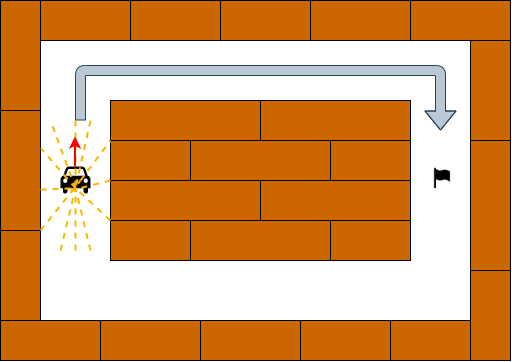}
	\caption{Autonomous driving course for \emph{F1/10} car}
	\label{fig:map}
\end{figure}

In safety-critical \acp{CPS}, timing correctness can be as important as functional correctness.
Consider the case published in \cite{ivanov2019case} where an autonomous \emph{F1/10} vehicle is using LiDAR to drive safely through a racetrack, as depicted in Figure~\ref{fig:map}.
A corner approaches.
The vehicle must react appropriately, ensuring that two properties are met: firstly, that the corner is detected and the steering is changed to avoid crashing; and secondly, that the decision to steer the car around the corner is completed in a timely fashion.
In other words, if the car controller takes too long to process the change in the road, the output of that controller is \emph{incorrect} and will potentially lead to a crash.

Unfortunately, while traditional control code for \ac{CPS} such as autonomous vehicles is designed to be amenable to both functional verification (e.g. ensure that the car will turn) and timing verification (e.g. ensure that this decision will take place quickly enough), often modern control systems involving \ac{AI} do not.
Instead, popular learning frameworks such as Keras~\cite{Keras} and Caffe~\cite{Caffe} rely on training/validation, simulation, and deployment testing, approaches which only provide weak guarantees of correctness~\cite{sann}.
In addition, approaches such as those provided through TensorFlow Lite~\cite{TensorFlowLite} are popularising the usage of these kinds of neural networks in embedded devices without further examination of the safety implications involved.

To address this there is a push in the literature to bring formal methodologies into the \ac{AI} domain~\cite{tripakis2018data,ASurveyOnMethodsForTheSfetyAssuranceOfMachineLearningBasedSystems}, especially within the usage of \acp{ANN} such as \acp{MLP} and \acp{CNN}.
For example, the case study in \cite{ivanov2019case} was verified using reachability analysis, whereby the sigmoid activation function inside the neurons of the \ac{MLP} controller were converted into a non-linear hybrid system before being analysed using a tool they called \textit{Flow*}~\cite{DBLP:journals/corr/abs-1811-01828,10.1007/978-3-642-39799-8_18}.
Despite this, the issue of timing verification of \ac{ANN}-based controllers has received scant attention~\cite{sann}.

In this work we seek to address this challenge by providing a new design pipeline for converting neural networks trained in Keras to time-predictable C code.
This generated code, once deployed upon a suitable time-predictable architecture, may then be statically analysed to formally derive timing bounds.

The rest of this paper is organised as follows.
Section~\ref{sec:lit} discusses the state of the art in this area.
Section~\ref{sec:compiler} discusses our solution to this challenge, where Keras neural networks are compiled to functionally equivalent C code.
Finally, Section~\ref{sec:results} presents our evaluation and Section~\ref{sec:conclusions} concludes.

\section{Background}
\label{sec:lit}

While it is sometimes common to consider \emph{real-time} execution as simply \emph{fast} execution, for safety-critical applications such as those found within \acp{CPS} it is not enough to simply execute your software quickly~\cite{roychoudhury2009embedded}, as rarely-executed branches not covered by testing may hide the presence of delays or other \emph{timing anomalies}~\cite{TimingAnomalies}. As such, for a system to be truly \emph{real-time}, it must be proven to meet its timing requirements via techniques such as \acf{STA}.
Unfortunately, while there has been plenty of work in the literature focussed on increasing the execution speed of neural networks (for instance the work presented in \cite{Redmon_2016_CVPR}, where Redmon et al. present a fast execution framework for \acp{CNN}), the issue of formal timing predictability for \acp{ANN} has received less attention.

In many cases, this is related to the underlying implementations of the neural network libraries used. 
For \ac{STA} of software to take place, two key requirements must be met~\cite{roychoudhury2009embedded}.
Firstly, it must be possible to convert program binaries into traversable control flow graphs.
This can be very difficult given the presence of program interpreters (for instance if the neural network is implemented in a language such as Python); the presence of a complex runtime and/or \acf{OS}; and/or the presence of dynamic control flows (for instance code generalised to run many neural networks rather than being specialised to one).
Secondly, the code must be targeted at a time-predictable processor so that execution times can be derived for the paths through the control flow graph.

As such, software implementations of neural networks can be very difficult to time. 
For instance, TensorFlow Lite~\cite{TensorFlowLite}, which is one of the most popular tools used by industry to embed neural networks in cyber-physical devices, features a runtime and requires Linux to operate --- and as such it is not amenable to formal \ac{STA}.

To address this challenge, in \cite{sann} Roop et al. proposed synchronous execution of neural networks, where they may be designed with synchronous programming languages such as Esterel and then executed on a time-predictable processor such as Patmos~\cite{patmos:rts2018}.
However, their approach was limited, and not all networks could be timed.

Other approaches, observing that neural networks can be compiled to hardware rather than software (such as in Haddoc2~\cite{Haddoc2}) focus on timing hardware implementations of neural networks~\cite{CompositionalApproachRealTimeMachineLearning}.
Other hardware-based approaches also exist, although these mainly focus on performance~\cite{EfficientProcessingOfDNNsSurvey} rather than predictability. 
In addition, deploying neural networks to reconfigurable hardware is much more complex process than that provided through software tools such as Keras.

\section{Keras2C: Our Design Approach}
\label{sec:compiler}

There is no fundamental reason that implementations of neural networks must be so difficult to time. 
At their heart, neural networks such as \acp{MLP} and \acp{CNN} are just different kinds of feed-forward mathematical functions.
In this section we detail how a neural network, trained in Keras~\cite{Keras} to drive the autonomous vehicle track in Figure~\ref{fig:map}, may be formally implemented in a time-predictable manner.

\subsection{Design Process}

\begin{figure}[b]
	\vspace{-5mm}
	\centering
	\includegraphics[width=0.40\textwidth]{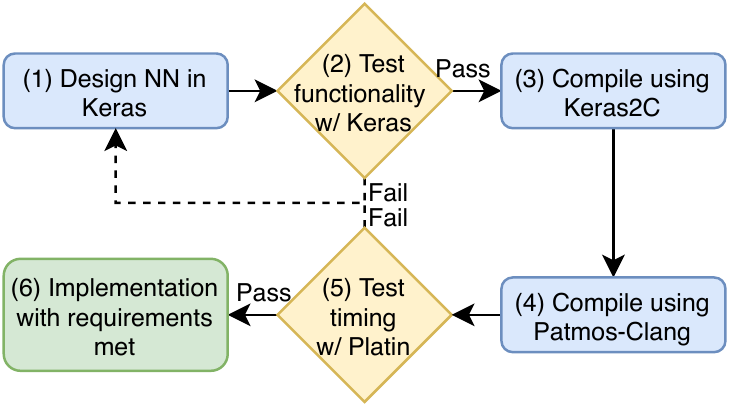}
	\caption{Keras2C Design Process}
	\label{fig:process}
	\vspace{-4mm}
\end{figure}

The Keras2C design process is depicted in Figure~\ref{fig:process}. 
Users will (1) design and train their networks using the Keras framework then (2) use the normal validation process to check functionality.
From (3) the saved Keras network files (an \sft{architecture.json} which describes the network structure, and a \sft{weights.h5} storing the network's trained weights) are given to the Keras2C tool which generates a folder of \sft{.c} and \sft{.h} files. 
These may then be (5) compiled with the open-source Patmos-clang compiler, and the output of the compilation further analysed with the existing \ac{STA} tool \emph{platin}~\cite{Platin} to derive \ac{WCET}.
If satisfactory, these files may then be considered the final implementation (6) of the neural network.

\subsection{Compiling \acfp{MLP}}

Let us examine this more formally for an \ac{MLP} type neural network, as this is suitable for driving the autonomous vehicle around the track in Figure~\ref{fig:map}.
An \ac{MLP} can be encoded as Definition~\ref{def:mdai:mlp}.

\begin{definition}
	\label{def:mdai:mlp}
	An \ac{MLP} can be formalised as a tuple $M = \langle I, O, L, N, \alpha, B, C, W, f \rangle$, where:
	\begin{itemize}
		\item $I$ is a finite collection of $n$ input variables with
		its domain being $\mathbf{I} = \mathbb{R}^n$
		\item  $O$ is a finite collection of $m$ output variables with
		its domain being $\mathbf{O} = \mathbb{R}^m$
		\item $L$ is a set of layers where $l_0$ represents the input layer and $l_{|L|-1}$ represents the output layer.
		\item  $N$ is a set of neurons.
		\item $\alpha: L \rightarrow N$ is a neuron mapping function $\alpha(l_n)$ where no neuron can appear in more than one layer. Neurons in the input layer are mapped to inputs $I$ (i.e. $|\alpha(l_0)| = |I|$), and neurons in the output layer are mapped to outputs $O$ (i.e. $|\alpha(l_{|L|-1})| = |O|$).
		\item $B: N \rightarrow \mathbb{R}$ is a function $B(n_i)$ which returns a real-valued bias for a given neuron.
		\item $C \subseteq N \times N$ is an ordered set of inter-layer connections $c_{i,j}$ between neurons $n_{i}$ and $n_{j}$ such that $n_i \in \alpha(l_k)$ and $n_j \in \alpha(l_{k+1})$ (i.e. connections can only go from a neuron in one layer to a neuron in the next layer).
		\item $W: C \rightarrow \mathbb{R}$ is a function $W(c_{i,j})$ which returns a real-valued weighting for a given connection $c_{i,j} \in C$.
		\item $f: \mathbb{R} \rightarrow \mathbb{R} $ is the neuron activation function (e.g. ReLU).
	\end{itemize}
	
	%$m \subset \eta \times L$ provides the mapping of neurons to layers. 
	%	We can describe the operation of $n$ as $n\left(I\right) = O$, i.e. $n$ describes a `black-box' untimed non-linear transformation function which converts inputs to outputs via the execution of each layer of neurons.
\end{definition} 

An \ac{MLP} according to Definition~\ref{def:mdai:mlp} can be straightforwardly executed using Algorithm~\ref{alg:mdai:ann:exe}.
Here, lines 1-3 set the input neuron layer to simply be their input values from $I$.
Lines 4-12 then compute the value of all other neurons (in the hidden and output layers), by first setting all other neurons  to their initial value i.e. their bias (line 6) and then summing their source neuron values multiplied by their connection weights (lines 7-9) before passing the sum into their activation function (line 10).
Finally, the output collection $O$ can be produced (lines 13-15).

\begin{algorithm}[t] 
	\caption{Execution of an \ac{MLP} $M$
		\label{alg:mdai:ann:exe}}
	\begin{algorithmic}[1]
		\small
		\ForAll {$n_i \in \alpha(l_0)$}
			\State $n_i \gets \sft{from\_input}(I, n_I)$
		\EndFor

		\ForEach {$l \in L \setminus l_0$}
			\ForEach {$n_i \in \alpha(l)$}
				\State $n_i \gets B(n_i)$
				\ForEach {$c_{h,i} \in C$ \textbf{from} $n_h$ \textbf{to} $n_i$}
					\State $n_i \gets n_i + n_h * W(c_{h,i})$
				\EndFor
				\State $n_i \gets f(n_i)$
			\EndFor
		\EndFor
		
		\ForAll {$n_i \in \alpha(l_k)$ \textbf{where} $k = |L|-1$}
			\State $\sft{set\_output}(O, n_i)$
		\EndFor
	\end{algorithmic}
\end{algorithm}

It is straightforward to see that if Algorithm~\ref{alg:mdai:ann:exe} is directly realised as static C code with bounded \sft{for} loops then this execution will be time-predictable.
However, while this approach is fundamentally the same as any used by the popular machine learning frameworks such as Keras, they focus on dynamic approaches for improving average-case execution time, featuring optimisations that can introduce unexpected delays and timing anomalies, and effectively prevent \ac{STA}.

In our case, we instead use the methodology in Algorithm~\ref{alg:mdai:ann:exe} to generate functionally equivalent C code from the saved Keras network files. 
The C code is static, only capable of executing the ANN it is generated from, and no operating system or runtimes are required.

\subsection{Other network types}

Definition~\ref{def:mdai:mlp} and Algorithm~\ref{alg:mdai:ann:exe} are designed for static execution of \acp{MLP}.
However, any stateless feedforward neural networks (such as \acp{CNN}) may be executed in a similar manner, as they too may be simply considered sequences of mathematical operations. 

\section{Evaluation}
\label{sec:results}

In this section, we evaluate the efficacy of our approach for a range of available benchmarks against two other methodologies.
Firstly, we examine our approach in a formal setting by benchmarking it against the existing results in \cite{sann} for time-predictable execution of neural networks on the Patmos architecture.
Then, we examine the raw scalability and raw performance of our approach more generally by benchmarking it against the popular TensorFlow Lite framework.

\subsection{Comparison with \acp{SANN}}

\cite{sann} introduces a synchronous framework for execution of \acp{ANN} that they term \ac{SANN}.
In this section we will benchmark our approach against their proposal.

\subsubsection{Methodology}
Using the \ac{SANN} approach, neural networks are trained offline and then compiled in a framework involving the Esterel programming language to C code.
However, while they could execute both \acp{CNN} and \acp{MLP}, only \acp{MLP} were compiled to predictable C code, which could then be executed over the time-predictable Patmos architecture, detailed here.

\subsubsection{The Patmos time-predictable processor}
\label{sec:patmos}

The Patmos~\cite{patmos:rts2018} processor is one part of the larger T-CREST~\cite{t-crest:2015} project,
which features a series of time-predictable hardware and associated tools, including interconnect, memories,
and software tool-chain including the LLVM-based compiler Patmos-clang
and the \ac{WCET} analysis tool Platin~\cite{compiler:platin:kps15}.

Patmos is a RISC style processor optimised for \ac{WCET} analysis.
An example of this optimization is the use of special cache types to aid \ac{WCET} analysis:
a stack cache, reserved for stack allocated data,
and a method cache for full function caching. If the
function does not fit into a cache block, it is broken into smaller functions
by the compiler.

\subsubsection{Benchmarking}

For this approach we compare four benchmarks from  \ac{SANN}~\cite{sann}.
Of the four, only XOR and Adder had formal analysis presented in the original paper, however our tool can generate predictable code for any \ac{MLP} and \ac{CNN} from Keras.
In addition, we also present the results for F1/10 running on Patmos, presented both with and without the fix16 library used in the \ac{SANN} benchmarks.
All benchmarks were analysed over a single-core 50MHz Patmos processor.

\begin{table}[h]
	\begin{tabular}{|l|l|l|l|}
		\hline
		\textbf{Benchmark} & \textbf{\begin{tabular}[c]{@{}l@{}}Number of \\ connections\end{tabular}} & \textbf{\begin{tabular}[c]{@{}l@{}}SANN WCET\\ (ms) \emph{from~\cite{sann}}\end{tabular}} & \textbf{\begin{tabular}[c]{@{}l@{}}Keras2C WCET\\ (ms)\end{tabular}} \\ \hline
		XOR                & 9                                                                         & 0.82                                                              & 0.1                                                                  \\ \hline
		Adder              & 15                                                                        & 0.49                                                              & 0.1                                                                  \\ \hline
		Rabbit             & 576                                                                       & Unavailable                                                                 & 2.93                                                                    \\ \hline
		Wolf               & 840                                                                       & Unavailable                                                                 & 4.03                                                                    \\ \hline
		F1/10 (fix16)      & 24,320                                                                       & N/A                                                                 & 82.83                                                                    \\ \hline
		F1/10 (float)      & 24,320                                                                       & N/A                                                                 & 3,235.39                                                                  \\ \hline
	\end{tabular}
	\caption{Benchmarking Keras2C against \ac{SANN}. Presented times are ``per invocation''.}
	\label{tbl:sann-v-keras2c}
\end{table}

As can be seen in Table~\ref{tbl:sann-v-keras2c}, the Keras2C code runs faster than the \ac{SANN} approach.
We believe this is due to overheads introduced by the Esterel compilation process.
Our approach is also more amenable to \ac{STA}, as the CNN library utilised in the original approach is not suitable for analysis. 
%If a system had multiple \acp{ANN}, max-plus type algebra like that used in \cite{sann} could be utilised to get compositional \ac{WCET} totals.
\subsubsection{Results for the F1/10 case study}

Consider the track depicted in Figure~\ref{fig:map}.
It is $20m$ by $10m$ with a width of $1.5m$.
The racing car starts from the mid-point of the left side of the track, drives forwards and makes two left turns, and will finally stop at the mid-point of the right slide of the track. 
%This takes $20s$.

The vehicle is initially stationary, and it accelerates until its velocity reaches $2.4ms^{-1}$.
It typically takes the car approximately $20s$ to complete the track.

\ignore{according to the following equation:
	\begin{equation}
	a = A_c \cdot M_c \cdot (T_c - H_c) - A_c \cdot V
	\end{equation}
	where,
	\begin{itemize}
		\item $a$ is the vehicle's acceleration,
		\item $A_c$ is the vehicle's acceleration constant ($= 1.633$),
		\item $M_c$ is the vehicle's motor constant ($= 0.2$),
		\item $T_c$ is the vehicle's throttle constant ($= 16$),
		\item $H_c$ is the vehicle's hysteresis constant ($= 4$),
		\item and $V$ represents the vehicle's current velocity.
	\end{itemize}
	As such, the vehicle will keep accelerating until its velocity reaches $0.2\times(16-4)=2.4m/s$.
}

In the original case study, the car's controller is expected to emit a control output every $100ms$ (i.e. every $0.24m$ of travel distance at top speed).
If this deadline is not met then an error will be introduced. If the deadline continues to fail to be met then further errors are introduced.
Eventually these errors may build up sufficiently that the car will crash.

Utilising our approach, the F1/10 case study is guaranteed to execute in $82.83ms$ when utilising the fix16 library for fixed point math.
However, if this library is not used, timing analysis becomes inaccurate due to the difficulties involved with predicting execution time for floating point computations~\cite{TowardsVerifiedConstantTimeFPCalcs}, and meeting the 100$ms$ deadline cannot be guaranteed.

\subsubsection{Analysing the LeNet-5 benchmark}

To demonstrate our ability to analyse \acp{CNN}, we also analyse the 60,000 parameter benchmark LeNet-5~\cite{LeNet-5}. Given the same Patmos core at 50MHz, the toolchain derives a \ac{WCET} of 751.84$ms$ (when using fix16) and 30,381$ms$ (when using floats).

\subsection{Performance Comparison with TensorFlow Lite}

TensorFlow Lite is a popular tool for executing neural networks on platforms such as smartphones and embedded computers (e.g. the ARM-based Raspberry Pi).
%As such, it is considered one tool for intelligent cyber-physical systems development.
It uses a dynamic runtime with a background garbage collector to execute its neural networks, and requires a full underlying \ac{OS}.
As such it is not amenable to \ac{STA}, unlike Keras2C.

\begin{table}[h]
	\begin{tabular}{|l|l|l|l|l|l|}
		\hline
		\multirow{2}{*}{\textbf{Benchmark}} & \multirow{2}{*}{\textbf{\begin{tabular}[c]{@{}l@{}}Number of \\ connections\end{tabular}}} & \multicolumn{2}{l|}{\textbf{TensorFlow Lite}}                                                                                 & \multicolumn{2}{l|}{\textbf{Keras2C}}                                                                                         \\ \cline{3-6} 
		&                                                                                            & \textit{\begin{tabular}[c]{@{}l@{}}Avg. \\ (ms)\end{tabular}} & \textit{\begin{tabular}[c]{@{}l@{}}WCET \\ (ms)\end{tabular}} & \textit{\begin{tabular}[c]{@{}l@{}}Avg. \\ (ms)\end{tabular}} & \textit{\begin{tabular}[c]{@{}l@{}}WCET \\ (ms)\end{tabular}} \\ \hline
		XOR                                 & 9                                                                                          & 0.027                                                         & 15.10                                                          & 0.0023                                                             & 0.06                                                          \\ \hline
		Adder                               & 15                                                                                         & 0.024                                                         & 10.06                                                          & 0.0026                                                             & 0.064                                                          \\ \hline
		Rabbit                              & 576                                                                                        & 0.046                                                         & 10.15                                                         & 0.026                                                             & 0.14                                                          \\ \hline
		Wolf                                & 840                                                                                        & 0.048                                                         & 13.07                                                          & 0.037                                                             & 0.13                                                          \\ \hline
		F1/10                               & 24320                                                                                      & 0.096                                                         & 2.269                                                        & 0.98                                                             & 1.401                                                          \\ \hline
	\end{tabular}
	\caption{Benchmarking Keras2C against TensorFlow Lite. Presented times are ``per invocation''. Lower is better.} 
	\label{tbl:tflite-v-keras2c}
\end{table}

Table~\ref{tbl:tflite-v-keras2c} presents our comparison with TensorFlow Lite on a Raspberry Pi Model 3B, where we measured the execution times of the network over a random input using the system clock. This is repeated for one million invocations.
For fairness, we omit time taken for the I/O of both frameworks as well as the first 100 ticks of TensorFlow Lite due to its significant ``wind up'' time.
As can be seen, while our approach is faster in the average case for the smaller networks, TensorFlow Lite outperforms our C code for the larger F1/10 network.
This is likely due to its internal optimisation routines which focus on average case performance.
However, these optimisations come at a cost, with unexpected delays in the worst case and overheads introduced for small networks.
As such, in all benchmarks our framework (as it is optimised for timing predictability) had a lower measured \ac{WCET}.

\section{Conclusions}
\label{sec:conclusions}

Neural networks are increasingly being adopted within \ac{CPS}.
However, while these systems might have safety-critical timing requirements, there are few approaches for verifying timing properties of neural network based code.
Through a series of benchmarks we have demonstrated our approach, Keras2C.
It addresses the challenge through a simple-to-use toolchain for converting neural networks trained in Keras to C code amenable to static timing analysis. 
Keras2C is made freely available at \emph{https://aitransformer.com}.
	
Future work could examine how certain optimisations could be integrated to improve average-case performance without compromising our timing predictability.

%Once these are obtained, combinations
%These can be then executed over a time-predictable architecture such as Patmos.

%\section*{Acknowledgment}

%The authors would like to thank...

% Can use something like this to put references on a page
% by themselves when using endfloat and the captionsoff option.
\ifCLASSOPTIONcaptionsoff
  \newpage
\fi

\bibliographystyle{IEEEtran}
% argument is your BibTeX string definitions and bibliography database(s)
\bibliography{main,IEEEabrv,hammond}

% that's all folks
\end{document}